\documentclass[sigconf]{aamas} 

\usepackage{hyperref}
\usepackage{url}

\usepackage{amsmath,amsfonts}
\usepackage{algorithm,algorithmic}
\usepackage{graphicx}
\usepackage{textcomp}
\usepackage{float}
\usepackage{booktabs}
\usepackage{bm}
\usepackage{threeparttable}
\usepackage{cleveref}
\usepackage{caption}
\usepackage{subcaption}
\usepackage{enumitem}
\setlist[description]{labelindent=.25cm,leftmargin=.5cm}
\setlist[itemize]{labelindent=.25cm,leftmargin=.5cm}

\usepackage{balance} 

\setcopyright{ifaamas}
\acmConference[AAMAS '23]{Proc.\@ of the 22nd International Conference
on Autonomous Agents and Multiagent Systems (AAMAS 2023)}{May 29 -- June 2, 2023}
{London, United Kingdom}{A.~Ricci, W.~Yeoh, N.~Agmon, B.~An (eds.)}
\copyrightyear{2023}
\acmYear{2023}
\acmDOI{}
\acmPrice{}
\acmISBN{}

\acmSubmissionID{918}

\title[AAMAS2023-Learning Rewards to Optimize Global Performance Metrics]{Learning Rewards to Optimize Global Performance Metrics in Deep Reinforcement Learning}

\author{Junqi Qian}
\affiliation{
  \institution{Shanghai Jiao Tong University}
  \city{Shanghai}
  \country{China}}
\email{kyle1994@sjtu.edu.cn}

\author{Paul Weng}
\affiliation{
  \institution{Shanghai Jiao Tong University}
  \city{Shanghai}
  \country{China}}
\email{paul.weng@sjtu.edu.cn}

\author{Chenmien Tan}
\affiliation{
  \institution{University of Nottingham Ningbo China}
  \city{Ningbo}
  \country{China}}
\email{chenmientan@outlook.com}

\begin{abstract}
When applying reinforcement learning (RL) to a new problem, reward engineering is a necessary, but often difficult and error-prone task a system designer has to face. 
To avoid this step, we propose \ours{}, a novel (deep) RL method that can optimize a global performance metric, which is supposed to be available as part of the problem description.
\ours{} alternates between two phases: (1) learning a (possibly vector) reward function used to fit the performance metric, and (2) training a policy to optimize an approximation of this performance metric based on the learned rewards.
Such RL training is not straightforward since both the reward function and the policy are trained using non-stationary data.
To overcome this issue, we propose several training tricks.
We demonstrate the efficiency of \ours{} on several domains.
Notably, \ours{} outperforms the winner of a recent autonomous driving competition organized at DAI'2020.
\end{abstract}


\newcommand{\BibTeX}{\rm B\kern-.05em{\sc i\kern-.025em b}\kern-.08em\TeX}

\newcommand{\St}{{\mathcal S}} 
\newcommand{\Ac}{{\mathcal A}} 
\newcommand{\T}{{\bm P}} 
\newcommand{\Tini}{{d_0}}
\newcommand{\Reward}{{r}} 
\newcommand{\Rw}{{\bm r}} 

\newcommand{\pf}{\rho} 
\newcommand{\pfparam}{\xi}
\newcommand{\pfp}{\hat\rho_\pfparam}

\newcommand{\hist}{\eta} 
\newcommand{\Hist}{{\mathbb H}} 
\newcommand{\ret}{\nu} 
\newcommand{\vret}{\bm\ret} 
\newcommand{\nO}{D} 

\newcommand{\vf}{{V}} 
\newcommand{\vvf}{{\bm V}} 
\newcommand{\Q}{{Q}} %
\newcommand{\vQ}{{\bm Q}} %

\newcommand{\Expect}{\mathbb E}
\newcommand{\Real}{\mathbb R}

\usepackage{etoolbox}
\newtoggle{final}
\toggletrue{final} 

\newcommand{\pw}[1]{\iftoggle{final}{#1}{{\color{orange} #1}}}
\newcommand{\jq}[1]{\iftoggle{final}{#1}{{\color{blue} #1}}}

\newcommand{\qjq}[1]{\iftoggle{final}{#1}{{\color{brown} #1}}}

\newcommand{\ours}{{\tt{LR4GPM}}}

\begin{document}


\pagestyle{fancy}
\fancyhead{}


\maketitle 



\section{Introduction}\label{sec:intro}

Reinforcement learning (RL) is a versatile machine learning technique to learn how to make good decisions by trial and error.
Although it has achieved excellent results in various domains \citep{SilverSchrittwieserSimonyanAntonoglouHuangGuezHubertBakerLaiBoltonChenLillicrapHuiSifreDriesscheGraepelHassabis17,FinnLevineAbbeel16,VinyalsBabuschkin19}, its application to real-world problems is still difficult due to various limitations of current RL techniques, e.g., large sample requirements, robustness of trained policies, or generalization issues.
In this paper, we focus on two specific related limitations, difficulty of reward engineering and insufficiency of the reward hypothesis, which we discuss next.

Reward engineering is one of the first steps a system designer has to go through when trying to solve a new control problem using RL.
It consists in designing a reward function, which serves two roles: (1) to guide the RL agent during training and (2) to define what is the correct behavior expected from a trained RL agent.
While sparse reward are easy to specify, learning from them may be unacceptably slow, especially on long horizon problems \citep{Ecoffet_2021}.
In contrast, dense rewards can accelerate RL training, but they may lead to undesirable behaviors \citep{RandlovAlstrom98,AmodeiOlahSteinhardtChristianoSchulmanMane16}.

The reward hypothesis \citep{Sutton2018} roughly states that intelligence may be formulated as the maximization of an expected accumulated rewards.
Although some argued for this conjecture \citep{Silver_Singh_Precup_Sutton_2021}, recent work \citep{Abel_Dabney_Harutyunyan_Ho_Littman_Precup_Singh_2021} shows some specific definition of tasks may not fit this hypothesis.
Actually, given the axiomatization of expected utility \citep{Machina88}, it is well-known in decision theory that some ordering over policies cannot be represented by it, e.g., when dealing with risk-sensitive decision-making \citep{Allais53}.

In this paper, we propose an RL method that works without the need of designing a reward function, but assumes instead that an overall performance metric is given to measure the quality of a policy.
This situation is actually common in practice.
For instance, in data center control, one may aim at optimizing power usage effectiveness \citep{Gao14}.
In network control, a classic objective function would combine both throughput and delay \citep{sivaraman2014experimental}.
In finance, one may consider the Sharpe ratio \citep{Sharpe94}, a performance metric that aggregates performance and risk.
In online advertising, one may focus on metrics like cost per action \citep{CPA08}.

As a first step in this direction, we assume that the form of the performance metric is known in this work.
We propose to automatically learn a reward function such that an approximation of the performance metric can be  optimized via the learned rewards by the RL agent.
The learned reward function may be vectorial to take into account the fact that the performance metric may be an aggregation of different objectives.

\paragraph{Main Contributions}
We propose a novel (deep) RL method that can directly optimize a given global performance metric without the need of a reward function.
We experimentally demonstrate its excellent performance over various domains (Mujoco, Iroko \citep{iroko}, SMARTS \citep{zhou_smarts_2021}) in various settings and various difficulty levels (standard RL performance metric to non-linear ones).
Notably, in the SMARTS simulator that was used in a competition\footnote{Organized during the \href{www.adai.ai/dai/2020/2020.html}{international conference on distributed artificial intelligence} (DAI) in 2020}, our method performs better than the winner and runner-up's submissions. 
Moreover, our experiments also provide some empirical evidence of the insufficiency of the reward hypothesis.

The remaining of the paper is organized as follows.
\Cref{sec:related} discusses the most related work.
\Cref{sec:background} recalls the necessary background and notations.
In \Cref{sec:problem}, we formally state the problem tackled in this paper and in \Cref{sec:method}, 
we detail our proposed method, \ours{}, 
discuss the difficulties of our problem, and
motivates the design of \ours{}.
\Cref{sec:experiments} presents the experimental results that validate our approach.
Finally, we conclude in \Cref{sec:conclusion}.

\section{Related Work} \label{sec:related}

The design of the reward function, which is related to the issue of value alignment \citep{ArnoldKasenbergScheutz17}, has been recognized as a difficult problem in the RL literature, as attested by the numerous publications in inverse RL \citep{Arora_Doshi_2021,Russell98} and behavior cloning \citep{ArgallChernovaVelosoBrowning09,Neurogammon}.
In a related direction, reward shaping \citep{NgHaradaRussell99}, some work \citep{El-AsriLarochePietquin12,Suay_Brys_Taylor_Chernova_2016} aims at learning a potential function to complement an existing sparse reward function.
Although various extensions \citep{ChristianoLeikeBrownMarticLeggAmodei17,IbarzLeikePohlenIrvingLeggAmodei18,Lee_Smith_Abbeel_2021} of theses techniques have been investigated, most work assumes that (often near expert) demonstrations are available or that an oracle is present to provide preferential feedback, which is not suitable in many domains.
In this paper, we do not make these assumptions.

Closer to our paper, some work proposes to learn a reward function, often online while training a policy.
\citet{Singh_where_2010} formulated the \emph{optimal reward design} problem where the goal is to learn a reward function that an agent would learn from such that it can maximize effectively the expected discounted reward computed with the system designer's reward function.
This setting is notably useful when the agent has some limitations (e.g., expressivity of parametric policy).
In this setting, \citet{Sorg_Lewis_Singh_2010} show that it may be worthwhile to learn a reward function, even if the designer's reward function is known, due notably to limitations of the policy representation.
\citet{Zheng_Oh_Singh_2018} learn intrinsic rewards for policy gradient methods.
In a related work, \citet{Agarwal_Liang_Schuurmans_Norouzi_2019} learn a reward function from sparse or underspecified rewards.
The multi-agent extension of optimal reward design has also been studied \citep{maord,Grunitzki_flexible}.
Our work can be interpreted as a generalization of optimal reward design \citep{Singh_where_2010} to more general performance metrics.
As a first step in this direction, we only consider the single-agent case in this paper.
We leave the extension of our approach to the multi-agent setting for future work.

Unlike all the previous methods that aim at maximizing the standard RL criterion, our approach can also deal with non-linear objective functions.
Thus, another related research direction regards extensions of RL to various considerations, e.g., risk-adversity \citep{TamarChowGhavamzadehMannor15,ChowTamarMannorPavone15},
fairness \citep{SiddiqueWengZimmer20,ZimmerGlanoisSiddiqueWeng21}, multi-objective settings \citep{RoijersSteckelmacherNowe18,Cheung2019,ReymondHayesRoijersSteckelmacherNowe21}, constraints \citep{ChowGhavamzadehJansonPavone16,AchiamHeldTamarAbbeel17}, or safety \citep{GarciaFernandez15,ZhangWeng21}.
In all these studies, the reward function is assumed to be given in contrast to our work.

\section{Background}\label{sec:background}

In this section, we recall the necessary background and notations about Markov decision process and (deep) reinforcement learning.


\subsection{Markov Decision Process}

A \emph{Markov Decision Process} (MDP) is a model for representing sequential decision-making problems under uncertainty. 
In such model, defined as a tuple $(\St, \Ac, \T, \Tini, \Reward)$, an agent repeatedly chooses some action $a$ from an action space $\Ac$ after observing a state $s$ from a state space $\St$.
The agent obtains an immediate reward $\Reward(s, a)$ after performing $a$ in $s$.
The transition function $\T$ defines the probability distribution $\T(s'\mid s, a)$ over next states and $\Tini$ is the initial state distribution.
The goal in this model is to determine how to select actions in states such as some performance metric is maximized.
We explain the details of the standard setting next.

The action selection is carried out via a \emph{(stochastic) policy} $\pi: \St \times \Ac \rightarrow [0,1]$, which is a mapping from state-action pairs to probabilities.
For convenience, we write $\pi(s, a)$ as $\pi(a \mid s)$, which specifies the probability of choosing an action $a$ in a state $s$.
When the probabilities of $\pi$ are either $0$ or $1$, it is said to be \emph{deterministic}.

We denote by $\hist = (s_0, a_0, \cdots, s_{T-1}, a_{T-1}, s_T)$ a $T$-step \emph{trajectory}, and by
$\Hist = \bigcup_n \St \times (\Ac \times \St)^n$ the set of all trajectories.
Under the standard RL criterion, the \emph{return} $\ret(\hist)$ of trajectory $\hist$ is defined by:
\begin{align}\label{eq: return}
    \ret(\hist) = \sum_{t} \gamma^t \Reward(s_t, a_t)
\end{align}
where $\gamma\in [0,1)$ is a discounted factor. 

The \emph{(state) value function} $\vf_{\pi}$ of a policy $\pi$ from an initial state $s$ is defined by:
\begin{align}\label{eqn:value function}
    \vf_{\pi}(s) = \Expect_{\pi, \T}[\ret(\hist) \mid s_0 = s]
\end{align}
%
Similarly, the \emph{action value function} $\vQ_{\pi}(s,a)$ is the expected return for applying from state $s$ and action $a$ followed by policy $\pi$:
\begin{align}\label{eqn: q function}
    \Q_{\pi}(s,a) = \Expect_{\pi, \T}[\ret(\hist) \mid s_0=s,a_0=a]
\end{align}

The usual goal in an MDP is to find an (optimal) policy $\pi^*$ that maximizes the \emph{expected discounted total reward} criterion:
\begin{equation}
    \Expect_\Tini[\vf_\pi(s)]
\end{equation} 
where the expectation is with respect to $s \sim \Tini$.
The value function of an optimal policy is said to be optimal.
In the RL setting, this goal is achieved without assuming the knowledge of the transition function or reward function.
In this paper, we refer to the expected discounted total reward criterion as the standard RL criterion.

\subsection{Standard Deep RL Methods} \label{sec: rl methods}
When the state space or action space becomes too large or continuous, parametric function approximation (e.g., neural network or linear function) is needed to allow generalization. 
In RL, the state (or action) value function, or the policy can be approximated by $\vf_\phi$ (or $\Q_\phi$), or $\pi_\theta$ where $\phi$ or $\theta$ denotes the parameters to be learned.

In value-based model-free RL, the optimal action value function is usually approximated. 
For instance, in DQN \citep{mnih_human-level_2015}, the parameters $\phi$ are learned by iteratively minimizing the squared difference between a regression target and the current approximation $\Q_\phi$:
\begin{align}
     \Expect \left(r+\gamma \max_{a'} \Q_{\phi^-}(s',a') - \Q_\phi(s,a) \right)^2 
\end{align}
where the expectation is with respect to the random sampling of tuples $(s,a,s',r)$, composed of a state, an action, a next state, and a reward respectively, from a replay buffer, and $\phi^-$ is a stabler copy of $\phi$.
The replay buffer stores the tuples $(s,a,s',r)$ experienced by the RL agent as it interacts with its environment.


In contrast to value-based methods, policy-based methods search for a good policy in a space of parametrized policies $\pi_\theta$.
\citet{sutton_policy_1999} proposes that  parameters $\theta$ should be updated in the direction of the following gradient $A_\pi(s,a) \nabla_\theta{\log \pi_\theta(a\mid s)}$ where $A_\pi(s,a) = Q_\pi(s,a) - V_\pi(s)$ is the \emph{advantage function}, which can be estimated in various manners.
In REINFORCE \citep{williams_simple_nodate}, the advantage function is estimated by the return from which a baseline is subtracted: $\hat{A}(s_t, a_t) = \ret_t - b_t(s_t)$, where $\ret_t$ is the return at time step $t$ and $b_t(s_t)$ is usually an estimated value function. 
In Proximal Policy Optimization (PPO) \citep{schulman_proximal_2017}, the advantage of $\pi_\theta$, approximated by $\hat{A}_\theta$, is estimated with $\lambda$-returns and a learned value function $\vf_\phi$.
PPO updates the parameters of $\pi_\theta$ in order to maximize the \emph{clipped surrogate objective function} defined as:
\begin{equation}
 J_{PPO}(\theta) = \Expect[\min(\omega_t(\theta)\hat{A}_{\overline\theta}(s_t, a_t), \mbox{clip}_\epsilon(\omega_t(\theta))\hat{A}_{\overline\theta}(s_t, a_t))]
\end{equation}
where $\omega_t(\theta) = \frac{\pi_{\theta}(a_t\mid s_t)}{\pi_{\overline\theta}(a_t\mid s_t)}$, $\overline\theta$ corresponds to the parameters of the policy that generated the training data, and $\mbox{clip}_\epsilon(x)$ clips $x$ in $[1-\epsilon, 1+\epsilon]$.
This objective function is defined as such to prevent large policy changes. 

\subsection{Vector Rewards}

In our proposed method, we allow the learned reward that is given to the RL agent after performing action $a$ in state $s$ to be a vector $\Rw(s,a) \in \mathbb R^\nO$ (with $\nO \ge 1$) instead of the usual scalar value $\Reward(s, a)$. 
The return defined in \Cref{eq: return}, the (state) value function defined \Cref{eqn:value function}, and the action value function \Cref{eqn: q function} can naturally be extended to the vector setting.
To make this clear, the vector counterparts are denoted in bold, $\vret$, $\vvf$, and $\vQ$.
Note that in contrast to multi-objective RL, we do not consider vector maximization.

\begin{table}[!tb]
    \centering
    \caption{Instances of our general framework}
    \label{tab:instances}
    \begin{tabular}{lll}
    \toprule
        Framework & $\rho$ & $\vret(\hist)$ \\
    \midrule 
      Standard RL & identity fun. & $\sum_t \gamma^t \Reward(s_t, a_t)$\\
      Fairness \citep{SiddiqueWengZimmer20} & fair function & $\sum_t \gamma^t \Rw(s_t, a_t)$ \\
      Risk-sensitivity \citep{FeiYangChenWang21} & $\frac{1}{\beta}\log$ & $\exp(\beta \sum_t \gamma^t \Reward(s_t, a_t))$\\
      General \citep{Cheung2019} & concave fun. & $\sum_t \gamma^t \Rw(s_t, a_t)$\\
      General \citep{RoijersSteckelmacherNowe18} & identity fun. & $u(\sum_t \gamma^t \Rw(s_t, a_t))$\\
    \bottomrule
    \end{tabular}
\end{table}

\section{Problem Statement}\label{sec:problem}

In this paper, we aim at learning a policy that  maximizes a given performance metric, assumed to be of the following form:
\begin{align}\label{eq:pb}
\max_\pi \pf(\Expect_{\pi,\T}[\vret(\hist)])
\end{align}
where $\pi$ is a policy, $\pf : \Real^\nO \to \Real$ is an aggregation function, $\Expect_{\pi,\T}$ is an expectation with respect to the distribution over trajectories $\hist$ induced by policy $\pi$, and $\vret : \Hist \to \Real^\nO$ is a trajectory evaluation function.
The interpretation of the overall objective function is simple: histories are first evaluated via function $\vret$ possibly over several dimensions if $\nO>1$, then the expected evaluations are aggregated into one score via function $\pf$.
We refer to this overall objective function as \emph{(\jq{global}) 
performance metric}.

Note that in \Cref{eq:pb}, $\vret(\hist)$ does not have to decompose additively like in the standard RL setting.
Furthermore, we assume that $\pf$ is differentiable, but $\vret$ does not have to be. 
This provides more flexibility in formulating how to evaluate a trajectory.
For instance, in the simplest case, the evaluation can be Boolean indicating whether a trajectory was successful or not.
However, since the evaluation applies to the whole trajectory, it can integrate diverse information along the trajectory to form an overall assessment.

This framework is very general.
For instance, it can formalize some notions of fairness \citep{SiddiqueWengZimmer20}, or risk-sensitivity \citep{FeiYangChenWang21}. 
It also includes the general multi-objective framework with concave objective functions proposed by \cite{Cheung2019} or that based on Expected Utility in \cite{RoijersSteckelmacherNowe18}.
We summarize in \Cref{tab:instances} how $\rho$ and $\vret$ are defined in all these cases.
In that table, 
$u : \mathbb R^\nO \to \mathbb R$ refers to a utility function.
The interested readers can obtain more details in the respective references.
Note that our general framework covers many other cases.
Two other examples (networking and autonomous driving) are introduced in \Cref{sec:experiments} for our experimental validation.

Problem \eqref{eq:pb} is not straightforward to solve via standard RL methods, since no reward function is readily available and function $\pf$ may be non-linear.
In the next section, we explain our proposed approach.

\section{Proposed Method} \label{sec:method}

We present first the general structure of our algorithm and its motivation.
Then, we explain why we include several training tricks.
We finish with a discussion about some possible alternative approaches.

\subsection{Basic Algorithmic Scheme}

Facing Problem~\eqref{eq:pb}, one may think of different straightforward approaches.
The simplest one consists in resorting to a REINFORCE-like algorithm.
However, such method is not sample efficient, and its performance is subpar in our experiments.
This observation suggests that learning an approximation of the performance metric is needed for both sample efficiency and performance.

Therefore, we propose to solve Problem~\eqref{eq:pb} by alternating between \pw{two phases:} (i) approximating its objective function (i.e., performance metric) and (ii) learning a policy.
Intuitively, a good approximation of the performance metric is obtained to guide policy training, which can then be used to generate training data for improving the approximation of the performance metric around the current policy.
This process can be repeated until convergence to a hopefully good policy.

\qjq{Apart from our method \ours{},} we also discuss at the end of this section several natural approaches to approximate the performance metric.
However, we have not managed to obtain very promising performances with any of them.
We report some of those results in \Cref{sec:experiments}, since they are used as baselines.
Below, we provide a \qjq{detailed} description of our approach.

Assuming that $\pi$ is a fixed policy, \jq{phase} (i) \pw{corresponds to this first problem}:
\begin{align}\label{eq:approx_pb1}
    \min_{\pfparam,\psi} ~ (\pf(\Expect_{\pi,\T}[\vret(\hist)]) - \pfp(\Expect_{\pi,\T}[\Rw_\psi(\hist)]))^2
\end{align}
where $\pfp$ is a parametric function (i.e., polynomial parametrized by $\pfparam$) and $\Rw_\psi(\hist)$ is the return of trajectory $\hist$ computed using a reward network $\Rw_\psi$ (i.e., neural network parametrized by $\psi$).

Learning a reward network $\Rw_\psi$ enables the usual temporal decomposition of the value 
of a trajectory, which helps \qjq{learn} a critic.
Moreover, it may allow to obtain a denser reward feedback that can better guide the RL agent during training.
Instead of using $\pf$ directly, an approximate $\pfp$ is learned to account for the approximation introduced by the reward network.
Moreover, learning both $\pfp$ and $\Rw_\psi$ can also be motivated and understood from the point of view of optimal reward design \citep{Singh_where_2010}.

If parameters $\pfparam$ and $\psi$ are updated to solve Problem~\eqref{eq:approx_pb1}, they would then form a good approximation of the objective function of Problem~\eqref{eq:pb} around trajectories generated by policy $\pi$.
However, since policy $\pi$ may be far from optimal, there is no reason to reach the best approximation possible of the performance metric around it.
Therefore, $\pfparam$ and $\psi$ only \jq{need} to be updated to improve the approximation so that they can be used to improve policy $\pi$.

Assuming now that $\pfparam$ and $\psi$ are fixed, \jq{phase} (ii) \pw{corresponds to this second problem}:
\begin{align}\label{eq:approx_pb2}
\max_\pi \pfp(\Expect_{\pi,\T}[\Rw_\psi(\hist)]) \,.
\end{align}
We assume this optimization problem to be on a space of parametrized policies $\pi_{\theta}$.
If we remove $\pfp$ in Problem~\eqref{eq:approx_pb2} and $\nO=1$, it would be a standard RL problem.
With $\pfp$, the problem is actually structurally similar to some recent extensions of RL to non-linear objective functions (e.g., \citep{Cheung2019,SiddiqueWengZimmer20}).

Again, if we were to update $\theta$ to solve Problem~\eqref{eq:approx_pb2}, we may obtain a good policy for the current reward function $\Rw_\psi$ (in combination with $\pfp$).
However, since $\pfparam$ and $\psi$ may not form a good approximation of the performance metric around a good policy for the original problem \eqref{eq:pb}, it is not necessary to solve Problem~\eqref{eq:approx_pb2} completely.
Therefore, $\theta$ only needs to be updated so that it can help improve the approximation produced by $\pfparam$ and $\psi$.
In the remaining, we drop subscript $\psi$ to avoid cluttering the notations since there is no risk of confusion because in our framework, the RL agent does not receive any other types of rewards.

\begin{algorithm}[!tb]
\caption{\ours{}}\label{alg:ours}
\begin{algorithmic}
\REQUIRE $\pf$ non-linear performance metric,
        $N$ number of iterations of the outer loop, 
        $m$ number of policies, 
        $n_B$ size of a minibatch, 
        $K$ number of the extra sets, 
        $n_\hist$ number of trajectories in a set,
        $n_H$ number of timesteps,
        $n_E$ number of trajectories.

\STATE Randomly initialize \pw{parameters} \jq{$\pfparam$ (performance metric),} $\theta$ (policy), $\phi$ (critic), and $\psi$ (reward)
\STATE Initialize $\psi^-$ (target network) as copy of $\psi$
\STATE Initialize replay buffer $\mathcal B$ of trajectories with trajectories generated by a uniformly-random policy
\FOR{$i=1, \ldots, N$}
    \STATE Sample $n_B+K$ sets of $n_\hist$ trajectories from $\mathcal B$
    \STATE Keep the top $n_B$ sets of trajectories w.r.t. $\pf$
    \STATE Update \jq{$\pfparam$} and $\psi$ using these $n_B$ sets of traj. to minimize  L$_2$ loss \eqref{eq:l2reward}
    
    \FOR{$j=1, \ldots, n_H$}
        \STATE Generate a set $E$ of trajectories with $\pi_\theta$ (the rewards are the weighted sum of values obtained with $\psi$ and $\psi^-$)
        \STATE Update $\theta$ (actor) with $E$ to optimize the extended $J_{PPO}$ \eqref{eq:ePPO}
        \STATE Update $\phi$ (critic) with the usual L$_2$ loss \eqref{eq:l2critic}
    \ENDFOR
    \STATE Generate $m$ sets of trajectories $E_1, \ldots, E_m$ with $\pi_\theta$
    \STATE Update $\theta$ (actor) to get $\theta_1$ to $\theta_m$ with each of these $E_1, \ldots, E_m$ to minimize the extended $J_{PPO}$ \eqref{eq:ePPO}
    \STATE Generate set $E'_k$ of trajectories with $\pi_{\theta_k}$ for $k \in \{1, 2, \ldots, m\}$
    \STATE Evaluate each $\pi_{\theta_k}$ with respect to $\pf$ using $E'_k$ 
    \STATE $\theta \gets \theta_k$ for the highest evaluated policy $\pi_{\theta_k}$
    \STATE Generate $n_E$ trajectories with $\theta$ and add them in $\mathcal B$ 
\ENDFOR
\end{algorithmic}
\end{algorithm}

More precisely, our algorithm (see~\Cref{alg:ours}), called \emph{Learning Rewards for  \jq{Global} Performance Metric} (\ours{}), consists of two nested loops.
In the outer loop, function $\pfp$ and the reward network $\Rw$ are trained such that the following loss $\mathcal L(\psi, \mathbb B)$ is reduced:
\begin{align}\label{eq:l2reward}
    \frac{1}{n_{\mathbb B}}
    \sum_{k=1}^{n_{\mathbb B}} \left(
    \pf\big(\frac{1}{|B_k|}\sum_{\hist \in B_k} \nu(\hist)\big) - \pfp\big(\frac{1}{|B_k|}\sum_{\hist \in B_k} \Rw(\hist)\big)
    \right)^2
\end{align}
where $\mathbb B = \{B_1, \ldots, B_{n_{\mathbb B}}\}$ is a mini-batch of set of histories sampled from a replay buffer $\mathcal B$.
To make the training of the reward network $\vret$ and the learning of $\pfp$ sample efficient, replay buffer $\mathcal B$ is filled with trajectories generated by the current policy.
This loss approximates the objective function of \eqref{eq:approx_pb1} via empirical means.
In addition, we perform an extra average to reduce variance.

In the inner loop, an RL agent is trained over a fixed number of steps to solve Problem~\eqref{eq:approx_pb2} assuming that $\pfp$ and $\Rw$ are fixed.
This problem, which is a variation of standard RL, can be solved efficiently \citep{SiddiqueWengZimmer20}.
For concreteness, we instantiate our proposition with PPO, but other RL algorithms would be possible.

Therefore, following the design of the PPO algorithm, the actor update is formulated such as it maximizes the following extended clipped surrogate objective function $J_{PPO}(\theta, E)$ defined as:
\begin{align}\label{eq:ePPO}
     \frac{1}{H\cdot|E|}\sum_{\hist \in E} \sum_{t=1}^{H}\min\big(\omega_t(\theta)\hat{A}_{\overline\theta}(s_t, a_t), \mbox{clip}_\epsilon(\omega_t(\theta))\hat{A}_{\overline\theta}(s_t, a_t)\big)
\end{align}
where $E$ is a batch of trajectories sampled by the current policy $\pi_{\overline\theta}$, 
$H$ is the length of the trajectories, 
$\omega_t(\theta) = \frac{\pi_{\theta}(a_t\mid s_t)}{\pi_{\overline\theta}(a_t\mid s_t)}$, $\mbox{clip}_\epsilon(x)$ clips $x$ in $[1-\epsilon, 1+\epsilon]$ and:
\begin{align}\label{eq:adv}
    \hat{A}_{\overline\theta}(s_t, a_t) = \pfp\left(\frac{1}{|E|} \sum_{\hist \in E} \sum_{k=t}^H \gamma^{k-t} \Rw_k^\hist\right) - \pfp\left(\vvf_\phi(s_t)\right)
\end{align}
corresponds to a generalized advantage function estimated for policy $\pi_{\overline \theta}$ that generated the trajectories in $E$. 
Note that the two terms in the advantage are transformed by $\pfp$.
Moreover, $\vvf_\phi$ is the critic learned in our extended PPO.
It is trained using a standard L$_2$ loss: 
\begin{align}\label{eq:l2critic}
    \frac{1}{H\cdot|E|}\sum_{\hist \in E}\sum_{t=1}^H \left(\sum_{k=t}^H \gamma^{k-t} \Rw_k^\hist - \vvf_\phi(s_t, a_t)\right)^2
\end{align}
where all the operations are performed componentwisely.
Recall both the return $\Rw$ and the state value function $\vvf_\phi$ take values in $\mathbb R^\nO$.
Therefore, the sum of the components of this critic loss is actually minimized at the end. 

Given \Cref{eq:ePPO}, 
the policy gradient $\nabla_{\theta} J(\pi_{\theta}, E)$ to update policy $\pi_\theta$ can be obtained easily via automatic differentiation when $\pi_\theta$ is represented as a neural network.
Note that the policy gradient used in our work is computed differently from that proposed by \citet{SiddiqueWengZimmer20}.
In their work, the non-linear aggregator ($\pf$ in our work) is applied on a clipped vector surrogate objective function, which is computed from a vector advantage function obtained from the difference of a vector Q-function and a vector value function.
While this is justified in fair optimization, such approach would not work in our setting, since applying $\pf$ on such advantages would not make sense.

We have now finished describing the basic algorithmic structure of \ours{}, which consists in repeatedly improving the approximation of the objective function in \eqref{eq:pb} and improving the current policy.
Note that both in the inner and outer loops, the parameters are updated several times using different randomly-sampled mini-batches, as it is common practice. 
Yet, this basic scheme is not sufficient, because of several difficulties, which we discuss next.

\subsection{Difficulties}

First of all, the environment for the policy learning is non-stationary since the reward function is updated dynamically in the outer loop. 
To alleviate the effects of the non-stationarity, we propose the following tricks that help stabilize the agent training process.
\begin{description}
    \item[Target reward network] To avoid the issues of sudden and large reward modifications in the policy learning, we keep a target reward network like in DQN \citep{mnih_human-level_2015}.
    The online reward network is copied at regular interval to the target one, which is used to train the RL agent.
    \item[Adaptive learning rates] 
    Since the reward network keeps changing, the policy is trained in a non-stationary environment.
    The usual approach in that case is to resort to constant learning rates.
    Moreover, following theoretical analysis \citep{KondaBorkar99} of actor-critic-like algorithms, we enforce the following constraints on the learning rates:
    the reward learning rate $\ge$ the critic learning rate $\ge$ the actor learning rate.
\end{description}

Without careful generation of the training data for the outer loop, the whole process can converge to a local optimum.
To alleviate this issue, we propose the following tricks:
\begin{description}
    \item[Optimistic policy update] For the last policy update in one inner loop iteration, we sample $m$ sets of trajectories $E_1, \ldots, E_m$ and update the current policy with each set to obtain $m$ new policies.
    Each of those policies is evaluated with respect to the performance metric by sampling from it a new set of trajectories.
    Only the best one is kept and a new set of trajectories, generated from it, is added to the replay buffer $\mathcal B$.
    This favors the current policy to be updated towards better policies and biases the data generation to include better trajectories.
    
    \item[Optimistic sampling] For approximating the performance metric, $n_B + K$ sets of mini-batches of trajectories are sampled and then evaluated with respect to the performance metric.
    Only the best $n_B$ mini-batches are kept and used for the update of the reward network.
    Here, $K$ is a hyperparameter controlling the optimism level.
    This biases the mini-batch sampling to include better trajectories and consequently improves the approximation around better trajectories.

    \item[Reward diversity] To differentiate the components of the vector reward network, we measure the distance/divergence over all pairs of returns $\vret(\hist)$ and maximize the sum of distance/divergences in the update of reward network. 
    We experimented with several possibilities, e.g., squared L$_2$, KL-divergence, JS-divergence, and Wasserstein distance (Recall that we can always normalize vector returns and treat them as probability distributions).
    We choose the squared L$_2$-distance because of its simplicity and its performance (see \Cref{sec:experiments}).
    Thus, the following additional loss, when averaged over several trajectories $\hist$, can then be combined with the loss in \eqref{eq:l2reward}:
    \begin{equation}
        \lambda_{RD} \sum_{i,j} \|\vret_i(\hist) - \vret_j(\hist)\|^2_{2}
    \end{equation}
    where $\lambda_{RD}$ is a hyperparameter controlling how much weight to put on this loss that promotes reward diversity.
 
\end{description}

Using these tricks in combination, \ours{} can solve Problem~\eqref{eq:pb} effectively and achieve excellent performance.


\subsection{Alternative Approaches}

In this subsection, we list a few alternative straightforward methods one could think of for approximating the performance metric:
\begin{itemize}
\item Following the reward hypothesis, one may aim at learning one scalar reward function to approximate the performance metric as an expected discounted reward.

\item Instead of learning a reward function, a Q-network could directly be trained to approximate the performance metric.
However, this requires the strong assumption that the environment can be reset in any state to train the Q-network in various states, since the performance metric in \eqref{eq:pb} needs to be computed as an average of many returns.

\item If the performance metric is an aggregation of several components, a vector reward function can be learned such that each of its components approximates one of those components (again with an expected discounted reward).
\end{itemize}
Such approximated performance metric can then be used to train a good policy in a similar scheme as in \ours{}.
However, these simple approaches either work poorly or are worse than \ours{} in our experiments, which show the difficulty of Problem \eqref{eq:pb}.











\section{Experimental Results}\label{sec:experiments}





We carried out several sets of experiments to validate our method\pw{, \ours{}}\footnote{To help reproduce our experiments, our code will be open-sourced after publication.}, against several baselines in three different domains: simple Mujoco control environments (\Cref{sec:mujoco}), a network control task (\Cref{sec:iroko}), and an autonomous driving task (\Cref{sec:smarts}).
We have different global performance metrics for different domains, from linear performance metric to non-linear performance metric that cannot be decomposed into direct sum of rewards.
We evaluated the policies obtained from \ours{} and baseline methods with respect to the global performance metric.
The baseline methods applied in each domain are listed in the\pw{ir respective} sections.

Hyperparameter optimization is performed in each domain for our method and the baseline methods, and each hyperparameter configuration is evaluated over 5 random seeds during the hyperparameter optimization.
The hyperparameter configuration corresponding to the best agent performance would then be rerun with  20 \pw{new} random seeds \pw{to compute an average result}.
\pw{Both during training and testing, a policy is evaluated with an empirical mean over 5 sets of trajectories, whose size is $n_\hist$ (see~\Cref{alg:ours}).}
For experiments in \Cref{sec:mujoco,sec:iroko}, simple grid search is applied for hyperparameter optimization.
For experiments in \Cref{sec:smarts}, Optuna~\citep{Akiba}, a hyperparameter optimization framework, is adopted to automatically find the hyperparameter configurations corresponding to best agent performance. 
\pw{Optuna is used because the baselines used in this domain are very strong.}
The hyperparameters used in our experiments are listed in the appendix.

\pw{In the next three subsections, we present our experiments in the three domains.
For each of them, we provide a short description of the environment, the definition of the global performance metric that is used, and finally the experimental results with some comments.
In all our figures, w}e plot the average value, standard error, and standard deviation of the performance measure for each method with solid line, dark shadowed area, and light shadowed area \pw{respectively}.
We also performed an ablation study in \Cref{sec:ablation} to understand the contributions of \pw{our} different tricks. 
More experiments can be found in the appendix.







\subsection{Mujoco Control Tasks} \label{sec:mujoco}



\pw{To check that our method can  work in the simplest setting (i.e., standard RL), we test it on some classic control tasks.
In this setting, although a reward function is defined, note that our \ours{} agent never sees the ground-truth rewards.
This reward function is only used to defined a performance metric (see below for more details).
}

\begin{description}[labelindent=0cm,leftmargin=0cm,parsep=.5em]

\item[Environment Description.]
We conduct our experiments in some MuJoCo \citep{mujoco} environments, specifically Hopper, Half-Cheetah, Rea{-}cher, and Walker2D, where in each environment, the agent with body parts connected by \pw{joints} are simulated in 2 
dimensions.
Observations provided by the simulator include the positions and (angle) velocities of the \pw{joints}.
The agent controls the torque applied on the \pw{joints} to move forward.
\pw{While the reward functions in Hopper, Half-Cheetah, and Walker-2D are dense, that in Reacher is sparse.}

\item[Global Performance Metric.]

In the MuJoCo control tasks, we simply adopt the discounted sum of \pw{the} rewards \pw{defined in those tasks} as the global performance metric.
This specific scenario illustrates a low-level case where the aggregation function $\rho$ is the \pw{identity} function and the trajectory evaluation function is \pw{additively decomposed}.
\pw{Therefore, here $\nO = n_\hist = 1$.}

\item[Experimental Results.]

\begin{figure*}[!tb]
    \centering
         \begin{subfigure}[b]{0.3\textwidth}
         \centering
         \includegraphics[width=\textwidth]{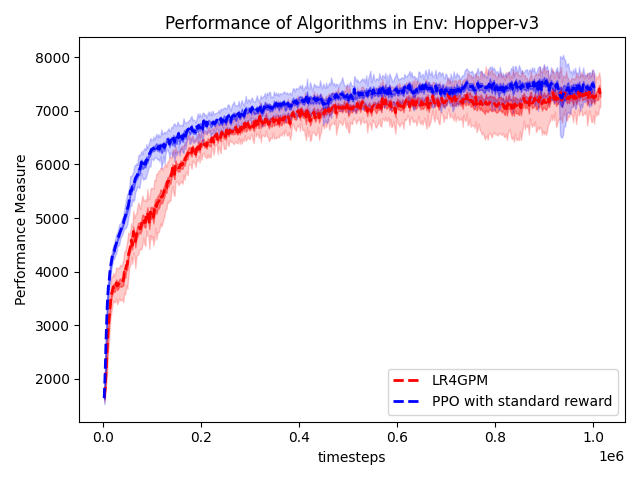}
         \caption{Hopper}
         \label{fig:hopper}
     \end{subfigure}
     \hfill
     \begin{subfigure}[b]{0.3\textwidth}
         \centering
         \includegraphics[width=\textwidth]{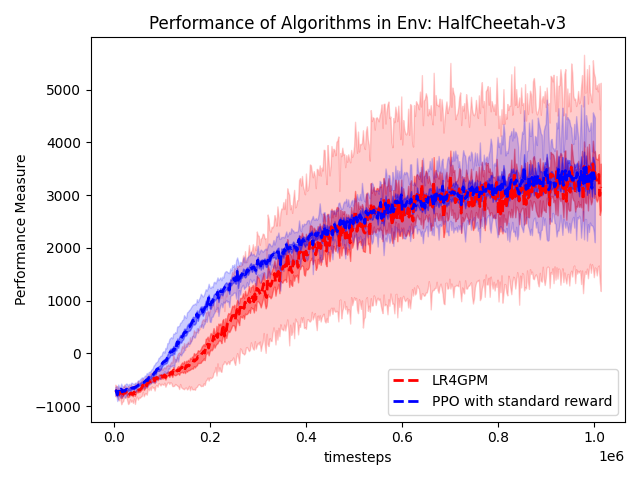}
         \caption{Half-Cheetah}
         \label{fig:halfCheetah}
     \end{subfigure}
     \hfill
     \begin{subfigure}[b]{0.3\textwidth}
         \centering
         \includegraphics[width=\textwidth]{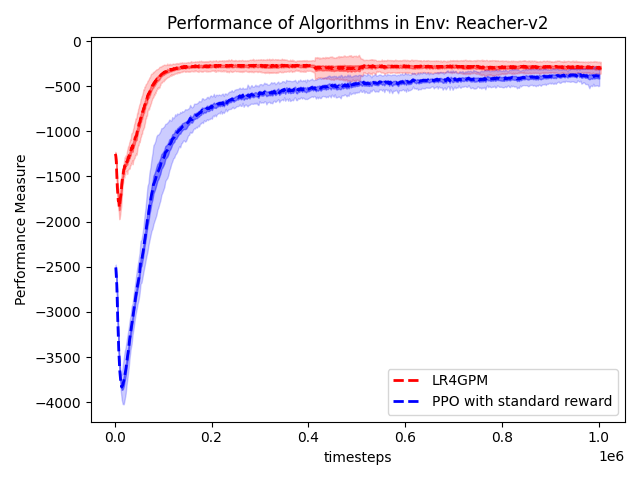}
         \caption{Reacher}
         \label{fig:Reacher}
     \end{subfigure}
     \hfill
     \begin{subfigure}[b]{0.3\textwidth}
         \centering
         \includegraphics[width=\textwidth]{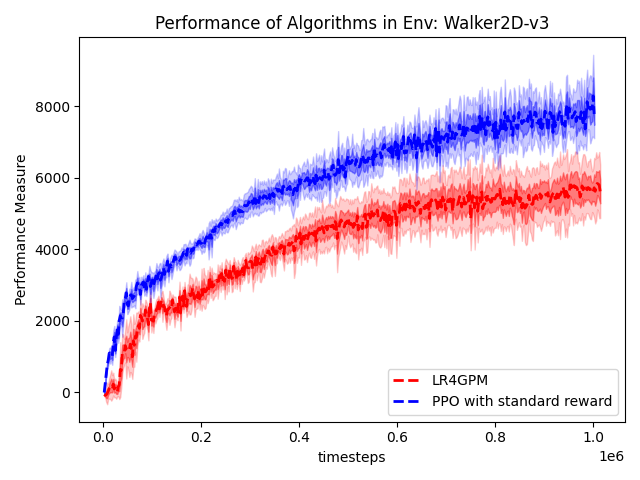}
         \caption{Walker2D}
         \label{fig:walker2d}
     \end{subfigure}
     \hfill
     \begin{subfigure}[b]{0.31\textwidth}
         \centering
         \includegraphics[width=\textwidth]{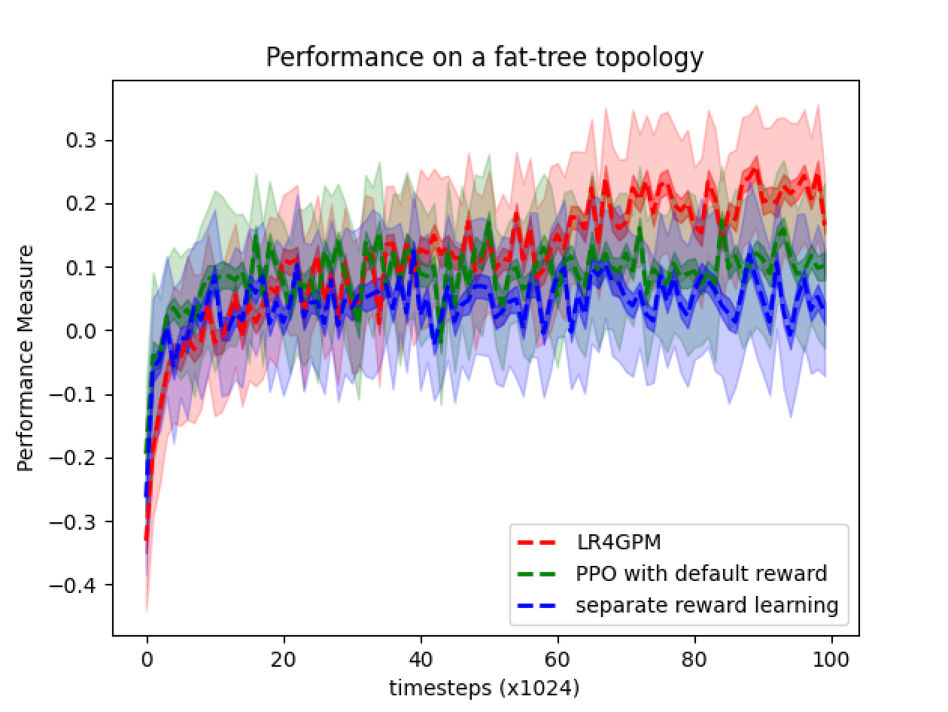}
         \caption{Iroko}
         \label{fig:iroko}
     \end{subfigure}
     \hfill
     \begin{subfigure}[b]{0.3\textwidth}
         \centering
         \includegraphics[width=\textwidth]{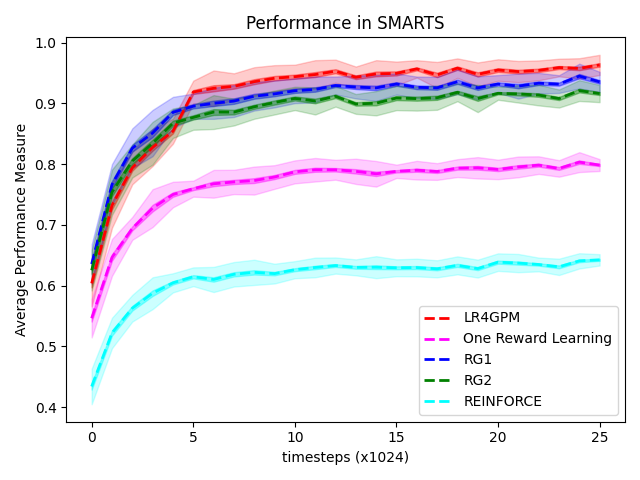}
         \caption{SMARTS}
         \label{fig:smarts}
     \end{subfigure}
    \caption{Algorithm performance. \qjq{\textmd{Performance measure of learned policy varied over timesteps. Mean and standard deviation are showed in the figure. Our method outperforms baselines in most cases.}} }
    \label{fig: alg performance}
\end{figure*}

\pw{In this setting, the most natural baseline is PPO \citep{schulman_proximal_2017}} trained with standard rewards.
We trained \pw{all agents} with 1 million steps.
The performance of \ours{} along with the standard PPO is illustrated in \qjq{\Cref{fig:hopper}, \Cref{fig:halfCheetah}, \Cref{fig:Reacher} and \Cref{fig:walker2d}}.
\ours{} can nearly match the performance of PPO in Hopper, Half-Cheetah, and Walker2D.
\pw{The biggest performance gap can be observed in Walker2D.}
Note that \ours{} solves a harder problem than PPO.
While PPO is guided by dense rewards, \ours{} needs to learn its own rewards.
Surprisingly, \ours{} achieves a better performance than the standard PPO in the environment of Reacher, indicating that our method is able to learn a\pw{n} effective and dense reward function from a global performance metric which is composed of sparse rewards.
\pw{To conclude, in dense reward tasks} \ours{} \pw{can achieve competitive} performance with little inferior convergence speed and optimality compared with \pw{the usual PPO.
In sparse reward tasks, \ours{} can be superior to PPO.
These points suggest} the potential of learning the near-optimal reward function from the global performance metric.
\end{description}


\subsection{Networking} \label{sec:iroko}

\pw{We perform experiments in this domain to demonstrate a practical use case of \ours{}.
In real world applications, a global performance is generally available, but a reward function is not.}

\begin{description}[labelindent=0cm,leftmargin=0cm,parsep=.5em]

\item[Environment Description.]
We conduct our experiments in the domain of data center control, where a centralized controller manages a computer network, \pw{which} is shared by hosts, in order to optimize the bandwidth of each host. 
\pw{For this problem, we use a Gym environment called Iroko \citep{iroko}, which is based on the Mininet \citep{mininet} network emulator, to simulate an RL environment for this data center control problem}.
The agent observes statistics from switches in the network and \pw{stores} them as a $d\times n$ \pw{matrix which contains traffic information going} through $n$ ports and $d$ network features. 
The agent controls the percentage of the maximum bandwidth each host \pw{is allocated}. 
In this domain, we focus on a \pw{relatively} complex scenario, which corresponds to a fat-tree topology.

\item[Global Performance Metric.]

\pw{In networking, it is common to evaluate a control by integrating the following two} aspects:
\begin{itemize}
    \item \textbf{Throughput}: summing the number of bytes received by all hosts divided by the delivery time (last receiving time - first sending time) for all packets.
    \item \textbf{Delay}: summing all the differences between the receiving time and the sending time divided by the number of packets.
\end{itemize}
\pw{We therefore define the global performance metric in this domain using the following equation} \citep{sivaraman2014experimental}:
\begin{equation}\label{eq:iroko_reward}
    \log(\mbox{throughput})-\delta \log(\mbox{delay})
\end{equation}
where $\delta$ is a weighted factor for the delay and it is set \pw{for simplicity} to be 1.0 in the experiments.

\item[Experimental Results.]

\pw{Note that when they propose Iroko, \citet{iroko} engineered a parametric reward function to achieve a good trade-off between throughput and delay.
Inspired by the work of \citet{sivaraman2014experimental}, they define it as follows:}
\begin{align}\label{eq:network reward}
    \sum_{i \in hosts} bw_i / bw_{max} - \omega \cdot (q_i / q_{max})^2 - \mbox{std}
\end{align}
where $bw_i$ (resp. $bw_{max}$) is the bandwidth of host $i$ (resp. max bandwidth), $q_i$ (resp. $q_{max}$) is the queue size of host $i$ (resp. max queue size), $\omega$ is a hyperparameter, and $std$ is the standard deviation of the bandwidth allocation (to favor fairness).
Following this reward function, the agent is \pw{therefore} encouraged to find a strategy to maximize the bandwidth usage as well as minimize the 
occurrence of queuing on the switch interface.
We trained the policies with 1 million steps for our method, \pw{PPO with} the default reward, and the separate component learning method introduced in \Cref{sec:method}.


The performance measure of the policies for different methods are shown in \Cref{fig:iroko}.
For PPO trained with the default reward, we show the agent with best performance among agents trained with different values for the factor $\omega$.
We see that separate reward learning works as well as PPO trained with the default reward, but \ours{} performs better than \pw{the} others \pw{in this networking problem, although it converges slightly slower. 
This is probably due to the fact that \ours{} needs some time to learn good rewards}. 
\end{description}

\subsection{Autonomous Driving} \label{sec:smarts}

\pw{We chose this third evaluation domain to fully demonstrate the potential of our approach.
This domain was proposed in a recent RL competition organized at DAI 2020.
In this competition, a global performance metric was defined to evaluate the submissions of the participants.
A default reward function was also provided, but it lead to a poor performance in terms of this performance metric.
Therefore, many participants spent a lot of effort to design a more efficient and informative reward function.
As we will show now, \ours{} can bypass the reward engineering step by learning a good reward function to train the RL agent.}

\begin{description}[labelindent=0cm,leftmargin=0cm,parsep=.5em]

\item[Environment Description.]
The last domain we conduct our experiments on is the SMARTS \citep{zhou_smarts_2021} (Scalable Multi-Agent RL Training School) platform which provides a simulation environment for reinforcement learning in autonomous driving. 
SMARTS offers a Gym interface  \citep{brockman_openai_2016} for RL agents.
We focus on the single agent situation in our experiments, where the agent is a specific moving vehicle in the scenario.
The observation space consists of the dynamic features of the moving vehicle and the road condition around the vehicle.
The agent can make a decision (accelerating, braking, turning) every step to determine the next vehicle state.

\item[Global Performance Metric.]

The performance of the agent \pw{is} evaluated according to the following global performance metric:
\begin{equation}
    \alpha \cdot \left( A \cdot \left( 1 - \frac{B}{B_{max}} \right) \right) + \beta \cdot \left( 1 - \frac{C}{W_{road}} \right) +\gamma \cdot \frac{D}{L_{route}}
\end{equation}
where $\alpha + \beta +\gamma = 1$ and $A$, $B$, $C$, and $D$ are defined as follows:
\begin{itemize}
    \item \textbf{Safety:} $A$ is the percentage of routes or missions completed without critical infractions (e.g., crashes with other vehicles). 
    
    \item \textbf{Time:} $B$ is the average time took to finish the task, also the timestep.
    
    \item \textbf{Control quality:} $C$ is the average distance to the center of the lane.
    
    \item \textbf{Valid distance travelled:} $D$ is the average driving distance along the predefined route.
\end{itemize}

The final score for a single map will be mapped into [0,1] by normalizing each independent metric. As shown in the formulation, $B_{max}$ represents the maximum time consumption; $W_{road}$ is half of the road width; $L_{route}$ is the length of the predefined route.
In this domain, we focus on the loop scenario where $\alpha$ is set to 0.4, $\beta$ is set to 0.1, and $\gamma$ is set to 0.5. 

\item[Baselines.]


We compare our approach with the following baselines in the experiments of autonomous driving:
\begin{description}
    \item[RG1, RG2:] \pw{PPO using} human-specified reward to optimize the global performance metric. \textbf{RG1} and \textbf{RG2} are the reward function \pw{engineered by} the winner and the runner up of the competition organized at DAI 2020.
    \pw{Those are therefore strong baselines.}
    \item[One Reward Learning:] \pw{PPO using} only one scalar reward to learn the global performance metric.
    \item[REINFORCE:] REINFORCE-like algorithm that directly optimizes the global performance metric in Problem~\eqref{eq:pb}.
\end{description}

\item[Experimental Results.]






\begin{figure*}[!tb]
    \centering
         \begin{subfigure}[b]{0.33\textwidth}
         \centering
         \includegraphics[width=\textwidth]{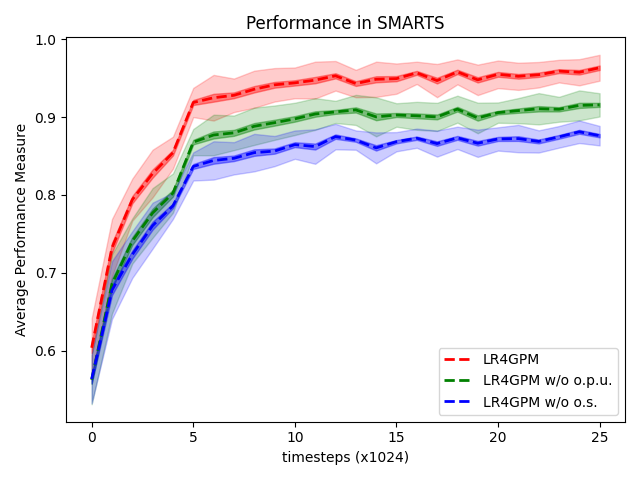}
         \caption{Optimistic update \& optimistic sampling}
         \label{fig:opu_os}
     \end{subfigure}
     \hfill
     \begin{subfigure}[b]{0.33\textwidth}
         \centering
         \includegraphics[width=\textwidth]{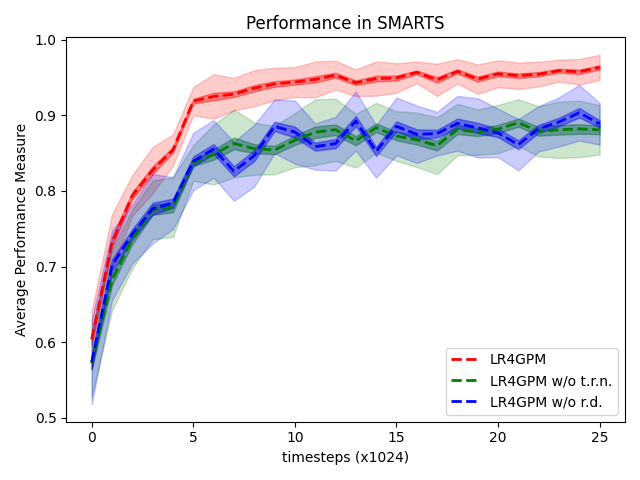}
         \caption{Reward diversity \& target reward network}
         \label{fig:rd_trn}
     \end{subfigure}
     \hfill
     \begin{subfigure}[b]{0.33\textwidth}
         \centering
         \includegraphics[width=\textwidth]{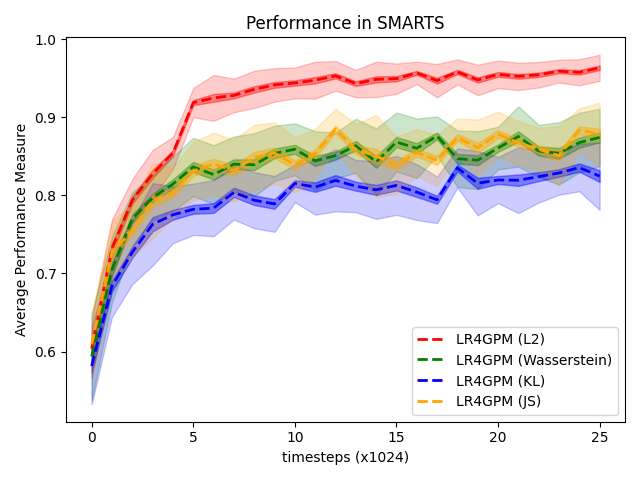}
         \caption{Different reward diversity measures}
         \label{fig:rd_type}
     \end{subfigure}
    \caption{Ablation study. \qjq{\textmd{Comparison of ablated training tricks in \ours{} performed in SMARTS platform shows their contributions. Mean and standard deviation are showed in the figure.}} }
    \label{fig: ablation study}
\end{figure*}

We present the results in \Cref{fig:smarts}.
We notice that \ours{} outperforms other methods in terms of performance measure.
The agent of \ours{} effectively learns from the rewards trained with the global performance metric.
\pw{The} two reward engineering methods also achieve good performance.
Learning with one scalar reward is worse than \ours{} because the non-linear performance metric cannot be directly decomposed into sum of rewards.
The REINFORCE-like algorithm also fails to achieve good performance. 


\end{description}

\subsection{Ablation Study} \label{sec:ablation}


In our method, we apply the following tricks in reward learning and agent training:
optimistic policy update,
optimistic sampling,
reward diversity,
and target reward network.
We perform an analysis of their contributions in the performance of our \ours{}.

\begin{description}[labelindent=0cm,leftmargin=0cm,parsep=.5em]
\item[Optimistic Policy Update.]
After we update the policy for the last time in the agent training (inner loop), multiple minibatches are sampled from the replay buffer to update the same number of candidate policies. 
The updated policy would be evaluated with respect to the global performance metric. 
We choose the one with best performance measure as the optimistic policy for the next reward update.
In order to disentangle the effects it has on \ours{}, we compare the performance of the agent trained with and without the optimistic policy update (green curve) in \Cref{fig:opu_os}.
We notice that it has little difference in the beginning and expand to some extent as the agent learns from the dynamically updated rewards.

\item[Optimistic sampling.]
When sampling the minibatch from the replay buffer, we sample much more than the size of the minibatch and then keep the best ones as a minibatch.
We test the effects of removing the optimistic sampling and sample the minibatch randomly from the replay buffer.
The blue curve in \Cref{fig:opu_os} shows that with the help of trajectory optimization, the agent can learn faster and achieve much better performance.

\item[Reward diversity.]
To alleviate the effects of non-stationarity of the dynamic environment, we add one more term in the objective function that increases the diversity of the different components of the reward function.
We choose the L$_2$ distance criterion to measure the differences between pairs of reward components.
\Cref{fig:rd_trn} shows that the introduction of reward diversity significantly helps in obtaining better performance and stabilizing the policy learning.

We also tried the following types of distance/divergence computation in the loss of reward learning: 
squared L$_2$ distance (applied in \ours{}),
Wasserstein distance,
KL divergence,
and JS divergence.
\Cref{fig:rd_type} shows their effects on the performance measure.
The agent achieves the best performance when applying L$_2$ distance in the loss formulation.
The agent trained with the loss where KL divergence is applied obtains the worst performance measure.
It could be highly related with the mismatch between the asymmetry property of KL divergence and the symmetry property in the reward vector.

\item[Target reward network]
In our method, we apply the moving average of the rewards in agent training to take into account the impact of the past reward from the target reward network, which helps to stabilize the policy learning.
We test the effects of removing the target reward network and directly applying the current rewards in the agent training.
\Cref{fig:rd_trn} shows that the performance measure of the agent decreases and the variance increases due to the lack of moving average rewards.
\end{description}

\section{Conclusion}\label{sec:conclusion}

In this paper, we introduce a new RL problem where the reward function is assumed not to be known, but the RL task is defined with a known overall performance metric.
For this problem, we design a novel algorithm that simultaneously (1) learns to approximate the performance metric with a (potentially vector) reward network and (2) trains a policy using this reward network.
Due to the difficulty of this problem, several tricks are called for to face the issues of non-stationarity training and early convergence to suboptimal policies.
Our proposition, \ours{}, is validated on several domains and shown to outperform various alternatives approaches.
Notably, \ours{} surpasses a reward engineered approach that wins in a recent RL competition, which is very strong baseline.

This indicates that \ours{} could be a promising approach to replace reward engineering and human expertise.
Indeed, our work can also be seen as a step towards automatic RL \citep{autoRL} where rewards, interpreted as hyperparameters, are adapted online.
Future work includes designing alternative algorithms for our new problem and extending the approach to the (cooperative or competitive) multi-agent setting.

\begin{acks}
This work is supported in part by the program of National Natural Science Foundation of China (No. 62176154) and the program of the Shanghai NSF (No. 19ZR1426700). 
\end{acks}


\bibliographystyle{ACM-Reference-Format} 
\balance
\bibliography{biblio_new,extra}

\clearpage
\appendix
\section{Appendix} \label{app:appendix}
\subsection{Detailed Environment Descriptions} \label{app:env}



\paragraph{Hopper}

The Hopper environment simulates a figure in 2 dimensions, where a torso is at the top, a thigh is in the middle, a leg is in the bottom, and a foot supports the body.
The goal is to apply torque on the three hinges that connect the four parts to make the figure move forward.
Observations are 11-dimensional, which indicate the position, angle, and (angular) velocity of the hinges.
Actions are 3-dimensional that control the torque applied on the hinges.
The reward is defined as the moving velocity minus the control cost, where the control cost is the square of the 2-norm of the action.
The agent receive additional bonus if it stays at a pre-defined healthy state.

\paragraph{HalfCheetah}

The HalfCheetah environment simulates a robot with 9 links and 8 joints in 2 dimensions.
The goal is to apply torque on the joints to make the robot run forward.
Observations are 17-dimensional, which indicate the (angular) velocities of the joints.
Actions are 6-dimensional, which control the torque applied on the joints except for the torso and head of the cheetah.
The reward is defined as the moving velocity of the cheetah minus the control cost, where the control cost is the square of the 2-norm of the action.

\paragraph{Reacher}

The Reacher environment simulates a robot arm with two joints.
The goal is to move the robot end effector, namely the fingertip, near by the target.
Observations are 11-dimensional, which indicates the positions of the slides and the angle (velocities) of the hinges.
Actions are 2-dimensional that control the torque applied on the hinges.
The reward is defined as the negative of the sum of the distance cost and the control cost, where the distance cost is the norm of the distance between the fingertip and the target and the control cost is the square of the 2-norm of the action.

\paragraph{Walker2D}

The Walker2D environment simulates a figure in 2 dimensions, where a single torso is at the top, two thighs are in the middle, two legs are in the bottom, and two feet support the body.
The goal is to apply torque on the six hinges that connect the six parts to make the figure move forward.
Observations are 17-dimensional, which indicate the position, angle, and (angular) velocity of the hinges.
Actions are 3-dimensional that control the torque applied on the hinges.
The reward is defined as the moving velocity minus the control cost, where the control cost is the square of the 2-norm of the action.
The agent receive additional bonus if it stays at a pre-defined healthy state.


\paragraph{Iroko}

\begin{figure}[!tbh]
    \centering
         \includegraphics[width=0.4\textwidth]{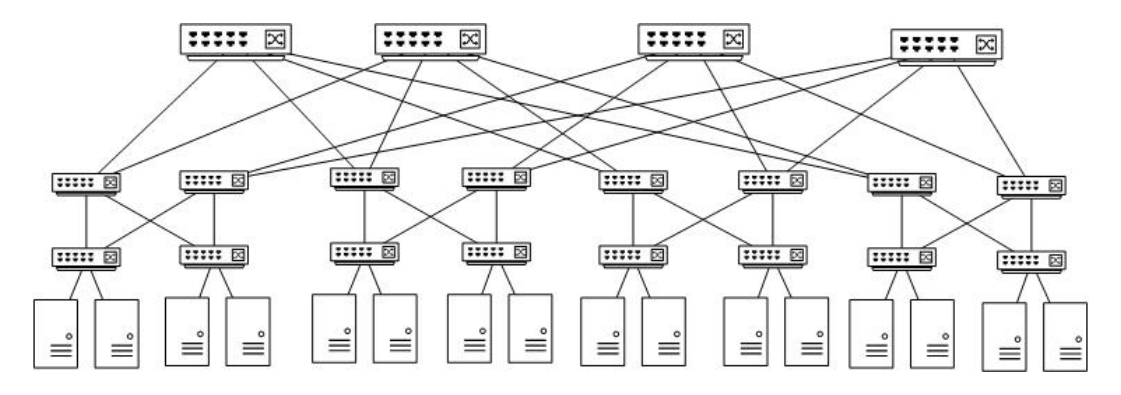}
         \caption{Scenario of fat-tree topology in Iroko \citep{iroko}}
         \label{fig:Iroko}
\end{figure}

Iroko simulates a controller that manages a network shared by multiple hosts (see \Cref{fig:Iroko}).
The goal is to allocate the bandwidth to hosts to maximize the utilization of the network.
Observations are $d\times n$ matrices of $d$ network features in $n$ ports, where the agents only have access to the switch buffer occupancy, the interface utilization, and the active flows.
Actions are $n$ dimensional, which control the percentage of the maximum bandwidth assigned to the host.
Rewards are consisted of three parts, namely the bandwidth reward, the queue penalty, and the dev penalty, where the bandwidth reward is the network utilization, the queue penalty is the queue size of switch ports weighted by the number of interfaces, and the dev penalty is the standard deviation of the actions.

\paragraph{SMARTS}

\begin{figure}[!tbh]
    \centering
         \includegraphics[width=0.4\textwidth]{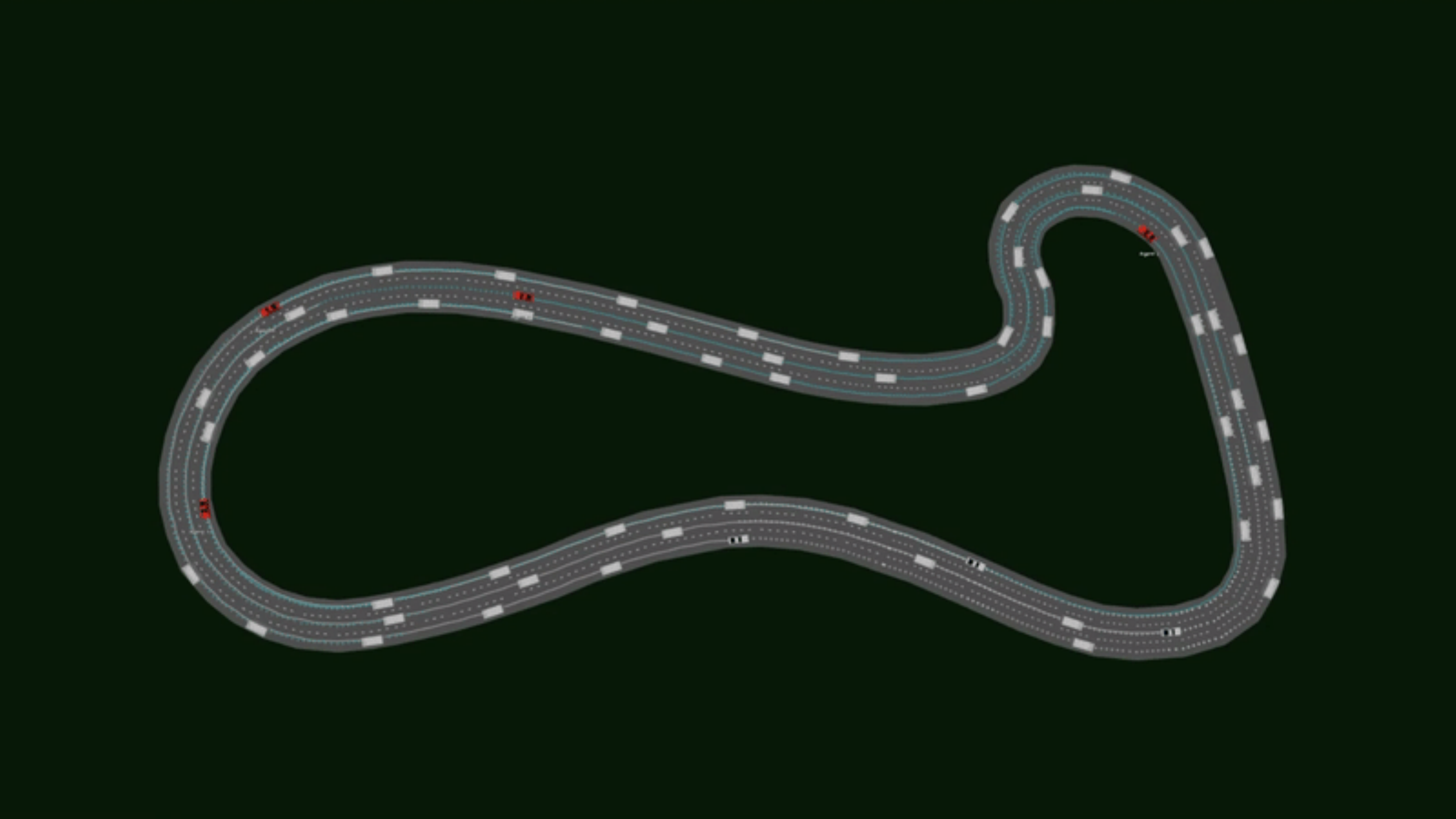}
         \caption{Loop scenario in SMARTS \citep{zhou_smarts_2021}}
         \label{fig:SMARTS}
\end{figure}

SMARTS simulates the driving interaction of multi-vehicles (see \Cref{fig:SMARTS}).
The goal of the ego-vehicle is to move along the mission road and avoid collisions.
The observation space is a customizable subset of multiple sensor types which include dynamic object list, bird view occupancy grid maps and RGB images, ego-vehicle states, and road structure ect.
The action space is determined by the chosen controller, where the types of continuous, actuator dynamic, trajectory tracking, and lane following are supported.
The reward is defined as the travelled distance along the mission road.
The agent configureably receive additional penalty for undesired behaviour, e.g., collision and off-road.

\subsection{Hyperparameters} \label{app:hyper}

\Cref{tab:hyp_ours} and \Cref{tab:hyp_ppo} summarize the hyperparameters for \ours{} and standard PPO in the experiments in \Cref{sec:experiments}.

\subsection{Additional Experiments}


\begin{figure}[!tb]
                \centering \includegraphics[width=0.4\textwidth]{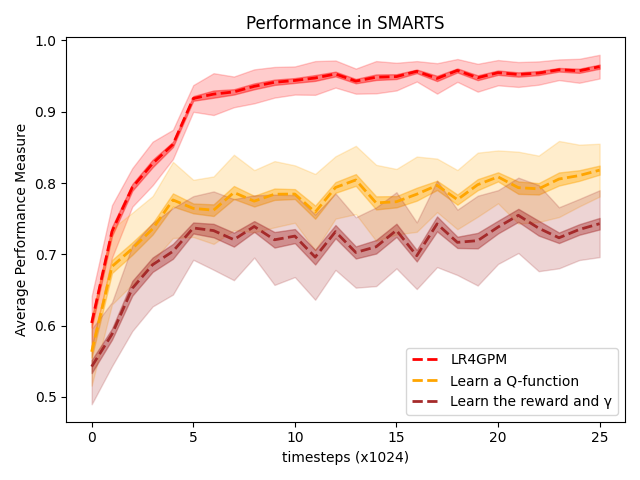}
                \caption{Algorithm performance for potential approaches}
                \label{fig:smarts_q_gamma}
\end{figure}

\paragraph{\pw{Alternative} approaches and results}

We also tried the following potential approaches to approximate the performance metric with RL:
\begin{description}
    \item[Learn a $Q$-function:] Learn a $Q$-function instead of a reward function to approximate the global performance metric by minimizing
    \begin{align}
        \left(Q^\pi(s, a) - \rho(\{\tau_j\})\right)^2
    \end{align}
    assuming all the $\tau_j$ starts from $(s, a)$ and are then generated by a policy $\pi$.

    \item[Learn a reward and $\gamma$:] Train the reward function with different episode lengths and different discount factors for each time step.
        The criterion to optimize the policy becomes:
        \begin{align}
            \mathbb E_\pi\left[\sum_t \gamma^t_t R_t\right]
        \end{align}
        where $\gamma_t$ is the discount factor for timestep $t$.
        Learning the reward function and the discount factors could be done by minimizing:
        \begin{align}
            \left( \frac{1}{N}\sum_i \sum_{t=0}^{|\tau_i|} \gamma^t_t R_\theta(s^i_t, a^i_t) - \rho(\lbrace \tau_j \rbrace_{j\in[N]}) \right)^2 
        \end{align}
        where $\tau_i = (s^i_t, a^i_t)_t$ and $|\tau_i|$ is the number of timesteps in episode $\tau_i$.
        The trainable parameters are $\theta$ and the $(\gamma_0, \ldots, \gamma_{H})$ where $H$ is the maximum length of an episode.
\end{description}
The corresponding experimental results are shown in \Cref{fig:smarts_q_gamma}. 
It is difficult to approximate a non-linear global performance metric either with a $Q$-function or with a learning reward discounted by the dynamic parameters.

\begin{figure}[!tbh]
                \centering \includegraphics[width=0.4\textwidth]{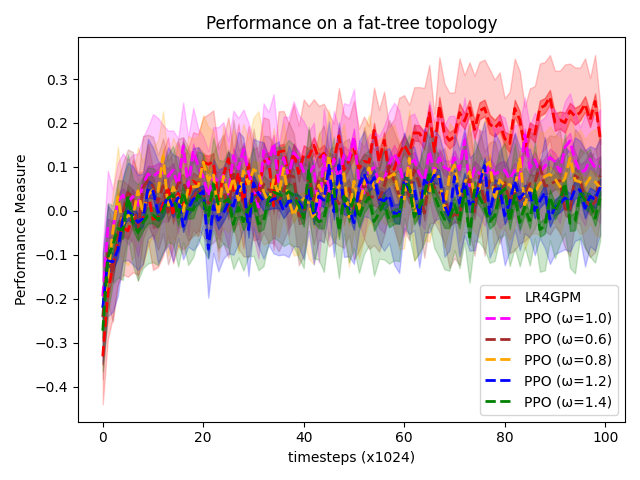}
                \caption{PPO with different weighted factor in the reward function}
                \label{fig:iroko_weights}
\end{figure}

\paragraph{PPO in Iroko} 
We have tried different weighted factor $\omega$ to train the agent with the default reward.
\Cref{fig:iroko_weights} shows that the agent trained with $\omega=1.0$ achieves slightly better performance than other values.
This is possible to have a strong relationship with the value of the weighted factor $\delta$ in \eqref{eq:iroko_reward}.

\begin{table*}[!tb]
    \centering
    \caption{Hyperparameters for \ours{} in the experiments}
    \label{tab:hyp_ours}
    \begin{tabular}{ccccccc}
    \toprule
     Reward Learning  & Hopper & HalfCheeatah & Reacher & Walker2D & Iroko & SMARTS\\
     \midrule
        $l_r$   & 0.008  &  0.004  &  0.008  &  0.004  &  0.003 &   0.0051   \\
        $\mbox{NET}_r$  &  $64^2$  &  $64^2$  &  $64^2$  &  $32^2$ & $64^2$  &  $256^2$    \\
        $n_B$   & 32  &  32  &  16  &  64  &  32  &  64  \\
        $n_{\mathcal{B}}$   &  64  &  128 & 64  &  64 &  1024 & 2048  \\
        $K$     & 10  &  10  &  5  &  10  &  10  &  20  \\
        $H$     & 32  &  64 &  32  &  64  &  32 &  50  \\
        $n_\eta$    &  5  &  5  &  5  &  5  &  10 &  20 \\
        $\lambda_{RD}$   &  /  &  / &  / &  / &  0.2 &  0.43 \\
        $m$    &  8  &  8  &  8  &  8  &  16 &   16 \\
        $n_E$  &  10 &  10 &  10 &  10 &  10 &   20 \\
     \midrule
     PPO      \\
     \midrule
        $l_a$   &   0.0008 &  0.0003 & 0.0008  & 0.0005   &  0.00008 &  0.0001 \\
        $l_c$   &   0.001 &   0.0005 & 0.001  &  0.0008  &   0.0008 &   0.0003 \\
        $\mbox{NET}_{PPO}$   &  $64^2$  &  $64^2$  &  $64^2$  &  $64^2$ &  $128^2$  &  $128^2$    \\
        $n_H$   &   2048 &  2048 &  1024  &  2048  &  1024 &  4096 \\
        $n_{epochs}$  & 10 &  10 &  10 & 10  &  5 &   5   \\
        $n_{B_{PPO}}$  & 128 &  64 &  32  & 64  &  64 &   128    \\
        $\gamma$    &  0.99  &  0.999  & 0.99  & 0.98  & 0.999  &  0.99 \\
        GAE-$\lambda$   & 0.95 &  0.95  &  0.98 &  0.8 & 0.98  &  0.99  \\
        $Cliprange$   &   0.2  &  0.2  &  0.1  &  0.2  &  0.2 &   0.4  \\
        $\lambda_{ent}$ &  0.001  &  0.000001 &  0.0001  &  0.001  &  0.0001 &   0.00001\\  
     \bottomrule
    \end{tabular}
\end{table*}


\begin{table*}[!tb]
    \centering
    \caption{Hyperparameters for standard PPO in the experiments}
    \label{tab:hyp_ppo}
    \begin{tabular}{cccccc}
        \toprule
        PPO hyperparameters  & Hopper & HalfCheeatah & Reacher & Walker2D & Iroko \\
        \midrule
        $lr$   & 0.0005  &  0.0003 & 0.0005  &  0.0003 &  0.0001  \\
        $\mbox{NET}_{PPO}$   &  $64^2$  &  $64^2$  & $64^2$   &  $32^2$  &   $128^2$    \\
        $n_H$   &  1024  & 2048  & 2048  &  2048 &  1024   \\
        $n_{epochs}$  &  10  &  10 &  10 &  10 &  5   \\
        $n_{B}$    &  64   &  64 &  64 & 128  &  64  \\
        $\gamma$    &  0.99  & 0.99  & 0.99  & 0.99  & 0.999   \\
        GAE-$\lambda$   &  0.99  & 0.99  & 0.999  & 0.99  &  0.95  \\
        $Cliprange$   &    0.2  &  0.4 &  0.2 &  0.2 &  0.2  \\
        $\lambda_{ent}$ &  0.000001 & 0.001  &  0.001 & 0.000001  &  0.0001  \\  
        \bottomrule
    \end{tabular}
\end{table*}

\end{document}